\documentclass[pmlr,twocolumn,10pt]{jmlr}

\makeatletter
\@ifundefined{mlhtrack}{%
  \newif\ifmlhfindings \newif\ifmlhdemo
  \newif\ifmlhneedspmlr \mlhneedspmlrtrue
  \newif\ifmlhanonymous \mlhanonymoustrue
  \newif\ifmlhneedsstatements \mlhneedsstatementstrue
  \newcommand{\mlhtrackname}{Proceedings Track}
  
  \newcommand{\mlhtrack}[1]{}%
}{}
\makeatother

\mlhtrack{proceedings}

\newif\iffinal
\finaltrue

\makeatletter
\iffinal\else\renewcommand*{\@titlefoot}{}\fi
\makeatother

\iffinal
    \ifmlhneedspmlr
      \jmlryear{2026}
    \fi
    \ifmlhfindings \jmlrproceedings{}{ML4H 2026 - Findings Track}\fi
    \ifmlhdemo     \jmlrproceedings{}{ML4H 2026 - Demo Track}\fi
    \jmlrproceedings{}{Accepted to ML4H 2026 (Proceedings Track)}
\else
    \jmlrproceedings{}{Submitted to ML4H 2026: \mlhtrackname}
    \jmlrworkshop{Machine Learning for Health (ML4H) 2026}
\fi

\usepackage{booktabs}
\usepackage{graphicx}

\usepackage{microtype}
\usepackage{xurl} 
\usepackage[document]{ragged2e} 
\usepackage{tikz}
\usetikzlibrary{arrows.meta,positioning,fit,backgrounds}
\usepackage[switch]{lineno}

\newsavebox{\promptsbox}
\newenvironment{promptbox}[2][]
  {\par\medskip\noindent
   \begin{lrbox}{\promptsbox}\begin{minipage}{\dimexpr\linewidth-1.6em\relax}%
   \setlength{\parindent}{0pt}\setlength{\parskip}{0pt}%
   \vspace{2pt}\textbf{\small #2}%
   \if\relax\detokenize{#1}\relax\else\ \ {\normalfont\small\itshape #1}\fi\par
   \vspace{3pt}\textcolor{black!30}{\rule{\linewidth}{0.4pt}}\par\vspace{4pt}
   \ttfamily\small\raggedright\setlength{\parskip}{5pt plus 1pt}%
   \ignorespaces}
  {\vspace{2pt}\end{minipage}\end{lrbox}%
   \fcolorbox{black!30}{black!3}{\usebox{\promptsbox}}\par\medskip}
\newcommand{\pq}{\char34}
\newcommand{\pind}{\hspace*{2em}}
\newcommand{\method}{RadRef}

\title[RadRef]{RadRef: A Reference-Free Metric for Radiology Report Generation}

\author[Schwartz et al.]{\Name{Shelly Schwartz}, \Name{Alan Ropp}, \Name{Michael B. Sneider}, \Name{Nur Yildirim}, \Name{Cree M. Gaskin}, \Name{Chirag Agarwal}\\[3pt]\addr University of Virginia, Charlottesville, VA, USA}

\begin{document}

\maketitle

\begin{abstract}
\RaggedRight
AI-generated radiology reports are commonly evaluated against radiologist-authored references, limiting the applicability of standard reference-based metrics when no reference is available. We introduce \method{}, a reference-free metric that scores report claims against image-derived beliefs over a clinical finding vocabulary, weighted by relevance to the clinical indication. On ReXVal, \method{} achieves $73.1\%$ agreement on non-tied same-radiograph report pairs with radiologist rankings, with $61.8\%$ pair coverage and $64.3\%$ tie-inclusive agreement; non-tied agreement is close to GREEN ($73.9\%$) and RadFact ($72.6\%$). It is also competitive on false findings and has its clearest relative result on omissions (missed findings). We further evaluate $100$ model-generated and $100$ radiologist-authored reports from $100$ MIMIC-CXR studies, all independently annotated by two board-certified radiologists. On same-image omission rankings, \method{} attains Kendall's $\tau_b=.295$, against $.386$ between the two readers themselves. In separate paired comparisons across generated reports, no compared metric significantly outperforms \method{} on omissions. The same image-grounded representation provides a false-finding readout, \method{}-FF, with $\tau_b=.32$ on ReXVal and $.29$ in the reader study, without requiring a reference. Controlled perturbations confirm that \method{} depends on the radiograph, report, and clinical indication. \method{} enables image-grounded evaluation before a radiologist-authored reference exists, supporting quality monitoring of report-generation systems in clinical settings, and complements reference-based metrics when one is available.
\end{abstract}


\ifmlhneedsstatements
\paragraph*{Data and Code Availability}
MIMIC-CXR and ReXVal are available to credentialed PhysioNet users. A de-identified example appears in \figureref{fig:pipeline} and \appendixref{apd:worked}. The code repository for \method{} is \url{https://github.com/shellyganga/RadRef-ML4H}. Clinical reports and images are not included in the code release.

\paragraph*{Institutional Review Board (IRB)}
The University of Virginia Institutional Review Board for the Social and Behavioral Sciences (IRB-SBS) approved the protocol \emph{Radiologist Evaluation of AI-Generated Radiology Reports} (protocol \#8659).
\fi

\section{Introduction}
\label{sec:intro}

Existing radiology-report metrics primarily evaluate generated reports against
radiologist-authored references, using lexical, embedding-based, structured or LLM-based
comparisons (\sectionref{sec:related}). This supports retrospective benchmarking but not the
settings in which no reference exists, including candidate selection, training-time evaluation
and deployment-time monitoring. In clinical deployment this gap is structural: the radiologist's
report is exactly what the system is asked to draft, so no reference exists when draft
quality matters most. Reference-based comparison also has a blind spot: if both the
candidate and reference omit the same finding, text-to-text comparison has no evidence that it
was missed.

\method{} maps the radiograph and the candidate report to a shared finding vocabulary and scores
how well the report's claims agree with image-derived beliefs estimated by a vision--language
model (VLM), weighted by clinical relevance (\sectionref{sec:metric}). It requires only the
radiograph, the candidate report and the clinical indication, and can therefore be computed the
moment a report is generated.

On ReXVal \citep{yu2023evaluating}, \method{} is competitive with leading reference-based metrics, attaining $73.1\%$ combined agreement and $69.8\%$ false-finding agreement, with its clearest relative result on omissions. A two-radiologist evaluation on 100 additional studies shows the omission signal carries to model-generated reports, where no compared metric significantly outperforms \method{}; the same representation also provides a false-finding readout, \method{}-FF, that tracks the readers' false-finding annotations (\sectionref{sec:reader}). Controlled perturbations confirm dependence on its intended inputs, and experiments across probe models characterize the role of image-belief quality.

\paragraph{Contributions.}
\begin{enumerate}
  \item We introduce \method{}, a reference-free, image-grounded metric for evaluating generated
        radiology reports (\sectionref{sec:metric}).
  \item We show that \method{} achieves $73.1\%$ combined agreement with radiologist rankings, close to leading reference-based metrics, with particularly favorable results on omissions, and that its false-finding readout, \method{}-FF, provides complementary error-specific evaluation from the same image-grounded representation (\sectionref{sec:within}). In a two-radiologist evaluation of model-generated reports, no compared metric significantly outperforms \method{} on omissions (\sectionref{sec:reader}, \appendixref{apd:reader}).
  \item We verify \method{}'s dependence on its intended inputs (\sectionref{sec:grounding}),
        characterize sensitivity to the image-belief probe model, and evaluate transfer of the
        scoring formulation to an alternative SNOMED CT finding vocabulary
        (\sectionref{sec:further}).
\end{enumerate}

\section{Related Work}
\label{sec:related}

\paragraph{Reference-based report metrics.} Existing clinical report metrics compare
candidate and reference text at increasing levels of abstraction. Lexical-overlap metrics
such as BLEU \citep{papineni2002bleu} and ROUGE-L \citep{lin2004rouge} correlate poorly
with factual correctness in radiology \citep{yu2023evaluating}, and BERTScore
\citep{zhang2020bertscore} softens exact matching to embedding similarity. CheXbert-F1
\citep{smit2020chexbert} and RadGraph-F1 \citep{yu2023evaluating,jain2021radgraph} compare extracted labels
and entity graphs. The most recent family uses a language model as the comparator, checking
entailment or counting errors by clinical category: RadFact \citep{bannur2024maira}, GREEN
\citep{ostmeier2024green} and CRIMSON \citep{baharoon2026crimson}. RadCliQ
\citep{yu2023evaluating} composes several metrics into a single error-count predictor. A
parallel line improves interpretability while remaining reference-based: LLM-RadJudge
\citep{wang2024llmradjudge} distills LLM pairwise report comparisons into a lightweight
judge, CLEAR \citep{jiang2025clear} compares reports attribute by attribute over
expert-curated conditions, and RadReason \citep{li2025radreason} produces per-error-type
sub-scores with written rationales. ICARE \citep{dua2025icare} compares reports through agent-generated clinical questions and cross-report answers, while ReportQA \citep{shi2026reportqa} uses report-based question answering to evaluate clinical information preservation.
These reference-based methods require a radiologist-authored reference, so they do not
score a report before a reference exists and cannot draw on image evidence absent from both texts.

\paragraph{Confidence in model-derived evidence.} Prior work studies uncertainty estimation and calibration in language and vision--language models \citep{farquhar2024semantic,tu2024calibration}. RadRef incorporates opposed probes, description evidence, and confidence weighting, and evaluates probe quality empirically.

\paragraph{Metric validation.} The standard validation is correlation with radiologist
error counts on a small annotated set, usually ReXVal \citep{yu2023evaluating}, which we
adopt here. Recent frameworks broaden this: ReEvalMed \citep{li2025reevalmed} probes
discrimination and robustness through controlled report edits, and RadEval
\citep{xu2025radeval} standardizes many metrics behind one interface with statistical
testing. We complement benchmark correlation with input-perturbation tests, study-independent-output
controls and a two-radiologist evaluation of model-generated reports.

\paragraph{Reference-free evaluation.} CLIPScore \citep{hessel2021clipscore} measures image--text embedding similarity. \citet{picha2026trust} combine visual grounding and text-guided image generation with clinical semantic comparison in a semi-reference-free radiology evaluation framework. Prior image-based fact-checking evaluates findings stated in generated reports \citep{mahmood2025factchecking}. RadRef additionally assesses omissions by evaluating an image-derived finding inventory beyond the report's assertions, weighted by clinical relevance, without a reference report at evaluation time.

\begin{figure*}[t]
\centering
\includegraphics[width=0.98\textwidth]{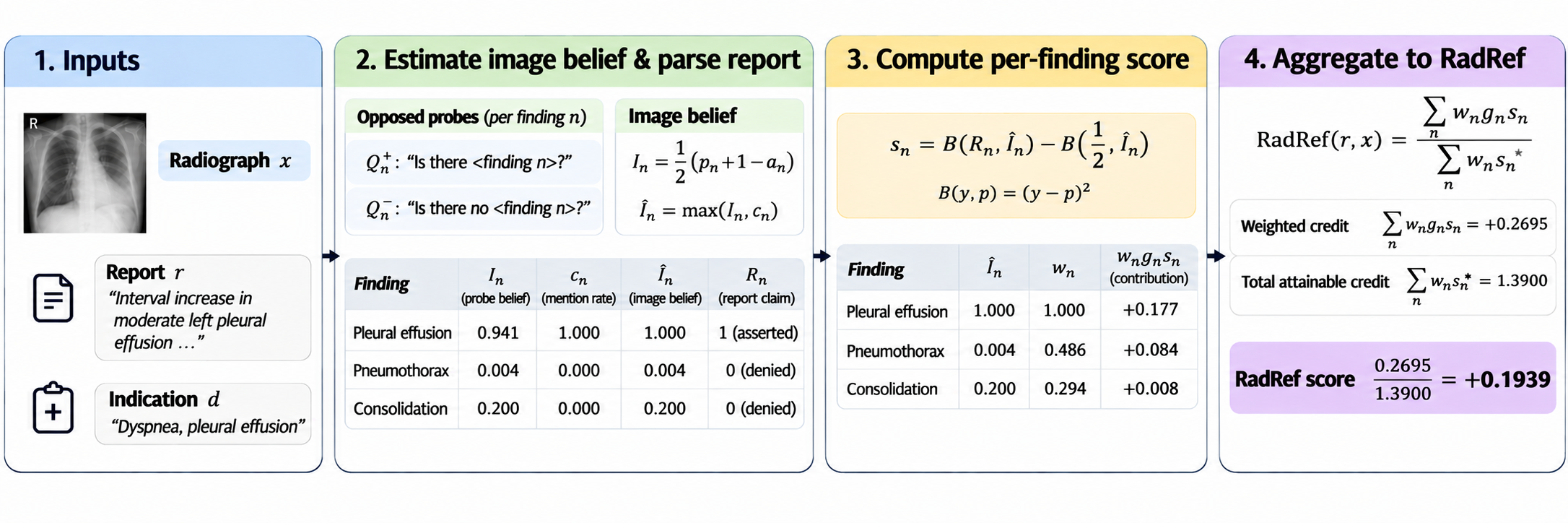}
\caption{RadRef worked example. The radiograph, report, and indication are mapped to finding-level image beliefs and report claims. The final score aggregates all retained findings, not only the displayed examples.}
\label{fig:pipeline}
\end{figure*}

\section{The RadRef Metric}
\label{sec:metric}

RadRef represents the radiograph and candidate report over the same finding vocabulary. A VLM
estimates the probability of each finding from the radiograph, and a language model maps the
report to asserted, denied or unmentioned findings. RadRef then scores agreement between the
report claims and image-derived beliefs, giving more weight to clinically relevant findings.
The result is a signed report-level score normalized for each radiograph. A report with no finding claims scores $0$. Positive values indicate net agreement with the image-derived beliefs, while negative values indicate net contradiction with those beliefs. Scores cannot exceed $1$, but there is no fixed lower bound.

\subsection{Finding vocabulary}
\label{sec:kb}

RadRef uses the 69-finding Chest ImaGenome vocabulary $\mathcal{K}$ \citep{wu2021imagenome},
collapsing synonyms during report parsing so that differently worded reports map to the same
entries (\appendixref{apd:prompts}). The scoring rule is not tied to this inventory: substituting a
62-term SNOMED CT vocabulary \citep{snomedct} changes combined ReXVal agreement by 3.8 points
(\appendixref{apd:kbsub}). Because the vocabulary is hierarchical, one abnormality can map to both a specific finding and
its broader parent terms. RadRef avoids double-counting by keeping the most specific supported
finding on each parent--child chain and removing its broader ancestors (\figureref{fig:dedup}).
Findings on separate branches remain distinct. This produces a de-duplicated set
$\mathcal{K}_x$ that is fixed for the radiograph $x$ and shared by every candidate report $r$ evaluated on that
study (\appendixref{apd:dedup}).

\begin{figure}[t]
\centering
\scriptsize
\begin{tikzpicture}[
  node distance=3mm,
  drop/.style={draw=black!30,fill=black!3,rounded corners=2pt,inner sep=3.5pt,minimum width=30mm},
  kept/.style={draw=green!45!black,fill=green!8,rounded corners=2pt,inner sep=3.5pt,minimum width=30mm},
  ar/.style={-{Latex[length=1.6mm]},draw=black!45}
]
\node[drop] (a1) {lung opacity};
\node[drop,below=of a1] (a2) {airspace opacity};
\node[kept,below=of a2] (a3) {alveolar hemorrhage};
\draw[ar] (a3) -- (a2);
\draw[ar] (a2) -- (a1);
\node[right=3mm of a1,black!50] {excluded};
\node[right=3mm of a2,black!50] {excluded};
\node[right=3mm of a3,green!40!black] {retained};
\draw[-{Latex[length=1.6mm]},draw=black!45]
  ([xshift=-5mm]a1.west) -- ([xshift=-5mm]a3.west)
  node[midway,rotate=90,above=1.5mm,black!55] {specificity increases};
\end{tikzpicture}
\caption{The most-specific rule. When multiple nested findings in the same hierarchy chain are
supported by the radiograph, only the most specific is retained in the scored set.}
\label{fig:dedup}
\end{figure}

\subsection{Beliefs, claims and the score}
\label{sec:score}

RadRef has three steps (\figureref{fig:pipeline}). First, a VLM estimates a belief
$I_n \in [0,1]$ for each finding from the radiograph. Second, a language model maps the
candidate report to a per-finding forecast: asserted, denied or unmentioned. Third, each report
claim is scored against the corresponding image-derived belief and the scores are combined across
findings. An unmentioned finding is the neutral action and contributes exactly $0$. We implement this comparison
with a Brier score in which leaving a finding unmentioned is the zero baseline.

\subsubsection{Image beliefs}
\label{sec:belief}
 We estimate each belief from two complementary signals: opposed yes/no probes and $M = 5$ free-form descriptions. The targeted probe model is the VLM that answers the finding-specific questions. For each finding $n \in \mathcal{K}$ we ask it both ``Is $n$ present?'' and ``Is the chest free of $n$?'' (\appendixref{apd:prompts}) and average the complementary ``yes'' probabilities into the probe-based belief $I_n = \tfrac12\left(p_n + 1 - a_n\right) \in [0,1]$, with $\tfrac12$ denoting maximal uncertainty. Opposed questions counter acquiescence: a model that tends to answer yes to both questions gives $p_n \approx a_n$, returning a belief near $\tfrac12$ rather than a confident error. We also use these probe probabilities to choose which hierarchy terms are scored. A finding is considered supported when $I_n \ge \tfrac12$. If both a specific finding and one of its broader parent terms are supported, we keep only the specific finding so that one abnormality is not counted twice. This choice fixes the finding set for the radiograph. We then add a second source of image evidence: RadVLM-7B \citep{deperrois2026radvlm} generates five free-form descriptions without being asked about any specific finding. Let $c_n$ be the fraction of the $M=5$ descriptions that mention finding $n$ (the number of descriptions mentioning $n$, divided by five). Thus $c_n\in\{0,0.2,0.4,0.6,0.8,1\}$. Repeated unprompted mentions provide additional support, so $c_n$ is used as a lower bound on the probe belief, giving
\begin{equation}
  \hat I_n \;=\; \max\left(I_n,\, c_n\right).
\label{eq:descfloor}
\end{equation}

\subsubsection{Report forecasts}
A language model maps the report onto $\mathcal{K}$ and labels each finding asserted, denied
or unmentioned (\appendixref{apd:prompts}). We use GPT-4o-mini
\citep{openai2024gpt4omini} at temperature $0$ for this structured extraction step. The parser can be swapped for another structured extractor without changing the scoring rule. Hedged assertions are treated as
present, while normality
statements such as ``lungs are clear'' are expanded into denials of the findings they exclude,
yielding the report forecast
\begin{equation}
  R_n =
  \begin{cases}
    1 & \text{the report asserts } n,\\
    0 & \text{the report denies } n,\\
    \tfrac12 & \text{the report does not mention } n.
  \end{cases}
\label{eq:report}
\end{equation}

\subsubsection{Per-finding score}
Each report claim is scored against the corresponding image-derived belief using the
negative Brier score \citep{brier1950verification},
$\mathrm{B}(q, I) = -[\,I(q-1)^2 + (1-I)q^2\,]$. We treat leaving a finding unmentioned as a neutral
baseline, represented by $q=\tfrac12$, and score each finding by how much the report
improves on that baseline:
\begin{equation}
  s_n \;=\; \mathrm{B}(R_n, \hat I_n) - \mathrm{B}(\tfrac12, \hat I_n).
\label{eq:skill}
\end{equation}
Substituting the three forecast values gives
\begin{equation*}
  s_n =
  \begin{cases}
    \hat I_n - \tfrac34 & R_n = 1,\\[1pt]
    \tfrac14 - \hat I_n & R_n = 0,\\[1pt]
    0 & R_n = \tfrac12,
  \end{cases}
\end{equation*}
An unmentioned finding therefore contributes exactly zero; an assertion earns credit only
when the belief exceeds $\tfrac34$, a denial only when it falls below $\tfrac14$, and
contradicted claims are negative. The baseline is a constant of the method, not estimated
from the data. Thus an omitted finding forgoes credit that a correct, radiograph-supported
assertion would earn rather than receiving a separate omission penalty. The denominator still includes the attainable credit for that supported finding, so omitting it reduces the report's normalized score relative to correctly asserting it.

\subsubsection{False-finding readout}
\label{sec:ff}
The same finding-level representation also supports an error-specific readout for false findings. We define \method{}-FF from assertions that directly conflict with the image-derived beliefs. An assertion contributes to \method{}-FF when $\hat I_n < \tfrac14$, the same threshold below which the \method{} score favors a denial:
\begin{equation}
  \text{\method{}-FF}(r,x) = -\sum_{n\in\mathcal{K}_x} \mathbf{1}\left[R_n=1 \land \hat I_n<\tfrac14\right].
\label{eq:radrefff}
\end{equation}
The negative sign preserves the higher-is-better direction of the main \method{} score: reports with fewer confidently contradicted assertions receive higher \method{}-FF values. \method{}-FF uses the same finding set, report parsing, and image-derived beliefs as \method{}; it is an error-specific readout of the same representation rather than a separate evaluator. The main \method{} score penalizes both unsupported assertions and incorrect denials. \method{}-FF isolates the unsupported-assertion component as a more specific measure of false findings.

\subsubsection{Aggregation}
RadRef gives more weight to clinically relevant findings that the belief model assesses
confidently, then normalizes by the maximum score attainable for that radiograph.

Each finding is weighted by its clinical relevance and the confidence of its image-derived belief,
\begin{equation}
  w_n = t_n^{2}\, \max\!\left(c_n,\, |2\hat I_n - 1|\right),
\label{eq:weight}
\end{equation}
 The tier $t_n \in \{1, 0.7, 0.5, 0.3, 0.1\}$ reflects relevance to the clinical indication (\appendixref{apd:prompts}), squared to emphasize directly relevant findings. We measure confidence in the image-derived belief as $|2\hat I_n - 1|$, the belief's distance from maximal uncertainty rescaled to $[0,1]$. These quantities measure the strength of the image-derived belief. Inconsistent opposed probes move beliefs toward $\tfrac12$. The mention rate $c_n$ (\sectionref{sec:belief}) preserves weight for findings that appear consistently in the descriptions even when the targeted probes are uncertain.

The gate $g_n = |\hat I_n - \tfrac12|^{\lambda}$, with $\lambda = \tfrac12$ in the primary configuration, shrinks a finding's contribution when the radiograph is uncertain about it. This prevents near-coin-flip beliefs from moving the score strongly. Each finding contributes $w_n g_n s_n$: $s_n$ gives the direction of credit, $w_n$ gives clinical importance, and $g_n$ controls how much uncertain radiograph evidence can matter.

RadRef sums these contributions and normalizes by a radiograph-specific scale, the maximum
per-finding credit before gating:
\begin{equation}
  \mathrm{RadRef}(r, x) \;=\; \frac{\sum_{n \in \mathcal{K}_x} w_n\, g_n\, s_n}
                                    {\sum_{n \in \mathcal{K}_x} w_n\, s^{\star}_n},
\label{eq:radref}
\end{equation}
where $s^{\star}_n = \max\!\left(\left|\hat I_n - \tfrac12\right| - \tfrac14,\ 0\right)$ is
the largest $s_n$ attainable for finding $n$: assert when the belief exceeds $\tfrac34$, deny
when it falls below $\tfrac14$, and otherwise leave it unmentioned. The denominator is the score of a
hypothetical best-claim report for that radiograph before the gate. Dividing by it prevents
radiographs with more confidently resolved findings from receiving systematically larger raw
scores. The gate is omitted from the denominator. If no finding contributes positive denominator mass, the score is undefined; such reports are omitted from analyses requiring that score (\appendixref{apd:undefined}). \appendixref{sec:gate-placement} discusses this choice and
\appendixref{apd:ablate} ablates each factor. \appendixref{apd:worked} gives a worked
example comparing the radiologist and MAIRA-2 reports for one case against the same image-derived beliefs.

\subsection{\method{} configurations}
\label{sec:settings}
All reported \method{} configurations use the same finding representation and scoring formulation. The primary RadRef configuration uses MedGemma-27B image beliefs without cohort-level calibration, allowing individual radiograph--report pairs to be evaluated independently. We additionally examine two alternative configurations, RadRef $+$calibration and RadRef $+$robust. \method{} $+$calibration applies finding-specific centering to account for persistent differences in the VLM's belief scale across findings, and \method{} $+$robust additionally places less weight on uncertain image evidence. All configurations remain reference-free: at scoring time they use the radiograph, the candidate report and the clinical indication, and never a radiologist-authored reference. Parameters are fixed a priori. Exact parameter settings and sensitivity analyses are in \appendixref{apd:variantparams} and \appendixref{apd:sensitivity}.

\paragraph{Recommended configuration.}
For MIMIC-CXR-style chest radiographs, we recommend the primary configuration: the Chest ImaGenome vocabulary, MedGemma-27B opposed image probes, five RadVLM-7B descriptions, and GPT-4o-mini report parsing. Use RadRef for overall findings and omissions, and RadRef-FF for unsupported assertions. The $+$calibration variant has higher ReXVal agreement but requires cohort-level belief statistics and is not the default for individual reports; $+$robust is primarily a cohort-monitoring sensitivity configuration. Vocabularies and probe models should be validated for the target clinical domain (\appendixref{apd:probe-diagnostics}).

\section{Data and Protocol}
\label{sec:setup}

\paragraph{ReXVal.} ReXVal \citep{yu2023evaluating} contains 50 MIMIC-CXR imaging studies
\citep{johnson2019mimic}, each with the radiologist's report and four
candidate reports, all 200 annotated by six board-certified radiologists. The candidates are
real reports of other patients, retrieved from the corpus to score highly under one of four
automatic metrics against the study's reference, rather than outputs of report generators. We evaluate the two error categories the finding
vocabulary represents, \emph{false prediction of finding} and \emph{omission of finding}, which we call \emph{false findings} and \emph{omissions}, and report their sum as
\emph{Combined}.

\paragraph{Additional MIMIC-CXR evaluation cohort.} ReXVal is small and contains retrieved reports rather than outputs from current generators, so we additionally evaluate on 500 MIMIC-CXR studies disjoint from ReXVal. To broaden clinical coverage, 200 frontal studies were selected to increase representation across 13 CheXpert pathology and device labels. Preference and indication analyses use a frozen subset of 498 studies; the archived 500-study export contains these studies and two additional replacements and is used for the score-contrast analysis. In the frozen subset, 466 probed images are PA/AP, 28 are lateral, and four have an unspecified view position. The cross-metric empty-report diagnostic uses a separate 150-study subset. Each study includes the radiologist's report and reports generated with the public MAIRA-2 \citep{bannur2024maira} and LLaVA-Rad \citep{chaves2024llavarad} models. We use this cohort for radiologist-preference diagnostics, empty-output controls and indication perturbations.

\paragraph{Models.} MedGemma-27B \citep{sellergren2025medgemma} supplies the image beliefs for the primary \method{} configuration; Lingshu-32B \citep{lasa2025lingshu} provides an alternative probe model for sensitivity analysis. A separate RadVLM-7B \citep{deperrois2026radvlm} generates the five shared free-form descriptions used with both probe models. GPT-4o-mini parses reports and assigns relevance tiers.

\subsection{Agreement with radiologist rankings}
\label{sec:gamma}

 For each ReXVal study we compare only reports for the same radiograph, holding case difficulty fixed. Pooling reports across studies confounds report quality with study-level differences, and the pooled radiologist error count correlates with report length alone at $\tau_b=+0.546$ (\appendixref{apd:pooled}). A pair is ranked correctly when the metric assigns the higher score to the report with fewer mean radiologist errors. \emph{Agreement} is the percentage of non-tied same-radiograph pairs ranked correctly. Kendall's $\tau_b$ provides the corresponding tie-aware rank statistic. Coverage, the fraction of pairs assigned distinct metric scores, is reported in \tableref{tab:dedup}.

\paragraph{Statistical analysis.} Confidence intervals and $p$-values use 2{,}000 study-level bootstrap resamples, preserving dependence among reports from the same study. Pairwise intervals and $p$-values are unadjusted for multiple comparisons and should be read as exploratory. The 498 evaluation-cohort studies come from 361 patients. Resampling by patient rather than study changes interval widths by factors of $0.91$ to $1.14$ and alters no conclusion.

{\setlength{\tabcolsep}{4pt}
\begin{table*}[t]
\floatconts
  {tab:within}
  {\caption{Same-radiograph agreement with ReXVal radiologist error rankings (50 studies, 200 reports). Agreement excludes metric ties. $\tau_b$ is tie-aware. $^\dagger$ marks values not significantly above chance ($p\ge .05$). \method{}-FF is reported only for false findings (\sectionref{sec:ff}). $^\ddagger$For \method{}-FF, $\tau_b$ is computed against false-finding counts; all other $\tau_b$ values use combined errors.}}
  {\footnotesize\begin{tabular}{lrrrr}
  \toprule
  & \multicolumn{3}{c}{Agreement, same radiograph} & \\
  \cmidrule(lr){2-4}
  Metric & False findings & Omissions & Combined & $\tau_b$\\
  \midrule
  GREEN                 & 71.4 & 68.5 & 73.9 & +0.399\\
  CRIMSON               & 77.7 & 68.7 & 76.9 & +0.480\\
  RadFact               & 76.9 & 55.8$^\dagger$ & 72.6 & +0.358\\
  RadGraph-F1           & 67.2 & 49.7$^\dagger$ & 65.3 & +0.284\\
  CheXbert-F1           & 56.2$^\dagger$ & 63.6 & 59.3 & +0.152\\
  CheXbert-emb          & 51.9$^\dagger$ & 62.1 & 57.3 & +0.137\\
  BERTScore             & 56.0$^\dagger$ & 55.3$^\dagger$ & 57.5 & +0.134\\
  ROUGE-L               & 46.1$^\dagger$ & 51.9$^\dagger$ & 50.2$^\dagger$ & +0.004\\
  BLEU-4                & 42.2$^\dagger$ & 49.1$^\dagger$ & 45.1$^\dagger$ & -0.093\\
  \midrule
  RadRef (MedGemma-27B)            & 69.8 & 70.2 & 73.1 & +0.353\\
  \quad $\hookrightarrow$ RadRef-FF & 86.7 & -- & -- & +0.324$^\ddagger$\\
  \quad + calibration              & 71.3 & 71.1 & 76.1 & +0.399\\
  \quad + robust                     & 66.7 & 65.8 & 70.1 & +0.308\\
  RadRef (Lingshu-32B)             & 71.8 & 69.3 & 68.6 & +0.290\\
  \quad + calibration              & 71.3 & 73.7 & 71.6 & +0.339\\
  \quad + robust                     & 69.9 & 72.0 & 69.5 & +0.306\\
  \bottomrule
  \end{tabular}}
\end{table*}}

\section{Results}
\label{sec:results}

\subsection{Agreement with radiologist error rankings}
\label{sec:within}

With MedGemma-27B probes, RadRef achieves $73.1\%$ combined agreement among non-tied report pairs (coverage $61.8\%$; tie-inclusive agreement $64.3\%$; $\tau_b=+0.353$), compared with $73.9\%$ for GREEN, $72.6\%$ for RadFact and $76.9\%$ for CRIMSON (\tableref{tab:within}). Paired tie-inclusive differences from these three leading baselines all have 95\% bootstrap intervals containing zero on the combined target; CRIMSON leads by 8.3 points under that convention (\tableref{tab:paired}). On false findings, its agreement is $69.8\%$, with differences from GREEN, RadFact and RadGraph-F1 not statistically distinguishable; only CRIMSON is significantly higher. The false-finding-specific \method{}-FF readout (\sectionref{sec:ff}) yields $\tau_b=.32$ on ReXVal, compared with $.29$ for the general \method{} score. On omissions, RadRef scores $70.2\%$, among the highest of the compared metrics, with paired intervals excluding zero against RadFact, RadGraph-F1 and BLEU-4; the differences from GREEN and CRIMSON on this target are not statistically distinguishable.

\subsection{Sensitivity to the radiograph, report and indication}
\label{sec:grounding}

\paragraph{Permutation.} To test whether RadRef depends on the radiograph rather than on
report-level regularities, we re-score each report against another patient's radiograph, so that no
report keeps its own image; a metric that ignored the image would be unchanged. Across 20
random re-pairings, observed agreement exceeds the radiograph-permuted distribution by
$2.0$--$4.1\sigma$ and the report-permuted distribution by $2.3$--$4.5\sigma$ across all six
configurations (\tableref{tab:placebo-main}). Pairings chosen to be maximally dissimilar rather
than random increase this margin (\appendixref{apd:permscheme}).

\begin{table}[!t]
\centering
\caption{Input permutations on \mbox{ReXVal}. Values are $\sigma$ above the mean across 20 trials; larger values indicate stronger dependence on the reassigned input. Only RadRef can be tested for radiograph dependence; a dash marks reference-based metrics, for which the radiograph is not an input.}
\label{tab:placebo-main}
\scriptsize
\setlength{\tabcolsep}{3pt}
\begin{tabular}{@{}lrr@{}}
\toprule
Metric & Radiograph & Report\\
\midrule
GREEN        & -- & $+4.8\sigma$\\
CRIMSON      & -- & $+4.3\sigma$\\
RadFact      & -- & $+6.0\sigma$\\
RadGraph-F1  & -- & $+3.5\sigma$\\
CheXbert-F1  & -- & $+1.3\sigma$\\
CheXbert-emb & -- & $+1.8\sigma$\\
BERTScore    & -- & $+1.9\sigma$\\
ROUGE-L      & -- & $+1.4\sigma$\\
BLEU-4       & -- & $-0.1\sigma$\\
\midrule
RadRef (MedGemma-27B) & $+4.1\sigma$ & $+3.6\sigma$\\
\quad + calibration & $+4.1\sigma$ & $+4.5\sigma$\\
\quad + robust & $+3.2\sigma$ & $+3.6\sigma$\\
RadRef (Lingshu-32B) & $+2.5\sigma$ & $+2.3\sigma$\\
\quad + calibration & $+2.0\sigma$ & $+2.9\sigma$\\
\quad + robust & $+2.3\sigma$ & $+2.3\sigma$\\
\bottomrule
\end{tabular}
\end{table}

\paragraph{Indication.} The indication reweights which findings matter but never alters the image-derived beliefs. The implemented evaluation uses the indication associated with each report; reports without a stated indication receive a generic chest evaluation. Thus, candidates for the same study can receive different relevance weights. Substituting the indication while holding the radiograph, report and image-derived beliefs fixed changes RadRef's rankings, confirming that the relevance weights affect the score (\tableref{tab:indication}). Full perturbation results are in \appendixref{apd:indctl}.

\label{sec:degen}\label{sec:separation}
\paragraph{Empty-output control.}
We also test an uninformative report containing no findings. The fraction of studies on which this empty report outscores MAIRA-2 varies widely across metrics despite similar ReXVal agreement, from $1.0\%$ to $90.7\%$ (\appendixref{apd:diagnostics}, \tableref{tab:empty-main}); this diagnostic is reported separately from the ranking results because score scales differ across metrics. For RadRef a defined empty report scores exactly zero by construction, so \tableref{tab:threedeltas} tests whether clinically meaningful reports fall on the expected side of that baseline: paired score differences favor radiologist-authored reports over either generator, while mean and median scores for all three report sources exceed the empty-report baseline, and all corresponding confidence intervals exclude zero. A component-swap analysis of radiologist reports scoring below zero is reported in \tableref{tab:below-empty-components}.

\begin{table}[t]
\centering
\caption{Across-image ranking agreement on 100 generated reports using protocol-audited counts. Columns name the target reader: target ties are excluded and predictor ties receive half credit. Inter-reader uses the other reader as predictor, giving directional denominators. $^{*}$95\% bootstrap CI excludes 50\%. RadRef-FF assesses unsupported assertions only.}
\label{tab:reader-main}
\footnotesize
\setlength{\tabcolsep}{3pt}
\begin{tabular}{@{}lrrrr@{}}
\toprule
& \multicolumn{2}{c}{Omissions} & \multicolumn{2}{c}{False findings}\\
\cmidrule(lr){2-3}\cmidrule(lr){4-5}
Metric & Reader 1 & Reader 2 & Reader 1 & Reader 2\\
\midrule
CRIMSON       & $63.1^{*}$ & $49.2$ & $75.8^{*}$ & $76.0^{*}$\\
RadFact       & $60.6^{*}$ & $57.0$ & $57.6$ & $58.0$\\
GREEN         & $61.0^{*}$ & $53.4$ & $51.6$ & $58.7$\\
RadGraph-F1   & $60.3^{*}$ & $50.4$ & $42.9$ & $57.7$\\
CheXbert-F1   & $64.2^{*}$ & $51.6$ & $48.5$ & $60.8$\\
CheXbert-emb  & $62.9^{*}$ & $48.0$ & $55.3$ & $63.2$\\
BERTScore     & $56.5$ & $50.2$ & $59.0$ & $67.7^{*}$\\
ROUGE-L       & $58.2$ & $53.0$ & $44.6$ & $65.7^{*}$\\
BLEU-4        & $63.4^{*}$ & $53.5$ & $47.4$ & $62.7^{*}$\\
\midrule
RadRef        & $64.1^{*}$ & $53.0$ & $50.1$ & $51.9$\\
\method{}-FF  & -- & -- & $53.4$ & $61.7^{*}$\\
\midrule
Inter-reader  & $65.6^{*}$ & $62.7^{*}$ & $71.0^{*}$ & $57.5^{*}$\\
\bottomrule
\end{tabular}
\end{table}

\subsection{Sensitivity to probe and vocabulary}
\label{sec:further}
\label{sec:oracletest}
The scoring formulation transfers to SNOMED CT, attaining $69.3\%$ combined ReXVal agreement versus $73.1\%$ with Chest ImaGenome; omission agreement is $55.7\%$ versus $70.2\%$ (\appendixref{apd:kbsub}). Keeping only the most specific supported finding avoids counting nested findings more than once (73.1\% versus 68.6\% when all hierarchy levels are scored; \appendixref{apd:dedup}). Probe choice produces larger changes in downstream agreement, motivating direct validation of candidate image models. We therefore assess candidate probes using finding AUC, confident false-positive and false-negative rates, confidence coverage, and held-out agreement with radiologist rankings (\appendixref{apd:probe-diagnostics}). Across eight independent sets of five descriptions per radiograph, combined agreement is $72.9\%\pm2.1$ percentage points (mean $\pm$ SD), compared with $73.1\%$ for the reference draw, showing stability to description resampling (\appendixref{apd:description-stability}). Component sensitivity is reported in \appendixref{apd:component-sensitivity}.

\subsection{Radiologist evaluation of model-generated reports}
\label{sec:reader}

We additionally evaluate reports generated directly from the radiograph rather than ReXVal's metric-selected retrieved reports. Two board-certified radiologists independently annotated false findings and omissions in 200 reports from 100 MIMIC-CXR studies. The set contains 50 MAIRA-2 and 50 LLaVA-Rad generations, each paired with the corresponding radiologist report. The readers were blinded to all metric scores.

We compare the radiologist and generated reports on each image using reader annotations, a comparison reference-based metrics cannot make when the radiologist report serves as their reference. We also compare metrics on the 100 generated reports separately for each reader (\tableref{tab:reader-main}). Using as-logged counts, RadRef orders the radiologist and generated reports in the same direction as the readers on $63.0\%$ of studies ($p=0.034$), rising to $68.5\%$ under the robust configuration ($p=0.002$). For omissions, the corresponding $\tau_b$ is $.295$ ($.10$--$.47$), compared with $.386$ ($.20$--$.55$) between readers (\tableref{tab:samerad}).

Across generated reports, no evaluated comparator significantly outperforms RadRef on omissions using as-logged counts (\appendixref{apd:reader}). With protocol-audited counts, RadRef's agreement with Reader 1 is $64.1\%$ (\tableref{tab:reader-main}). For false findings, RadRef-FF targets unsupported assertions (\sectionref{sec:ff}) and attains $\tau_b=.29$ ($.15$--$.42$) on same-image pairs, versus $.15$ for assertion count alone. Against mean reader error counts, its agreement is $61.1\%$ ($52.5$--$70.5$) on generated-report pairs.
\section{Discussion}
\label{sec:discussion}
RadRef evaluates report claims directly against image-derived evidence, enabling report evaluation without a radiologist-authored reference at scoring time. Its strongest relative result is on omissions. The image-derived finding inventory also allows RadRef to assess findings omitted from both candidate and reference. The reader study extends this result to model-generated reports. RadRef-FF provides a complementary readout for unsupported assertions from the same finding representation.

Probe selection is an important part of applying RadRef. The component analyses show that image-probe choice has a larger effect on agreement than resampling free-form descriptions. We evaluate candidate probes using finding-level discrimination, confident error rates, confidence coverage, and agreement with radiologist rankings (\appendixref{apd:probe-diagnostics}). This provides an empirical procedure for selecting the image-belief model for a given evaluation setting.

The scoring formulation also transfers to an alternative SNOMED CT finding inventory, with $69.3\%$ combined agreement versus $73.1\%$ for Chest ImaGenome; omission agreement is sensitive to finding terminology. Hierarchy de-duplication improves agreement by counting the most specific supported finding on each chain. Together, these results support adapting the finding vocabulary and validating the image probe for the intended domain. RadRef enables image-grounded evaluation of radiology reports without requiring a radiologist-authored reference, supporting both model evaluation and quality monitoring while complementing existing reference-based metrics.

\section{Limitations}
\label{sec:limitations}

RadRef scores finding-level content, not laterality, severity, temporal change, recommendations, or broader clinical completeness; out-of-vocabulary findings cannot be scored directly. Its image beliefs are model-dependent and not necessarily calibrated. The reader study is small, inter-reader agreement is imperfect (especially for false findings), and validation is confined to MIMIC-CXR, with possible training exposure. Independent image-level labels and external cohorts are needed to establish generalizability. Below-empty cases were not independently adjudicated.

\section{Acknowledgements}
We would like to thank all members of Aikyam Lab for their valuable feedback. S.S. is supported by the SDS Pilot Grant. C.A. is supported, in part, by grants from Capital One, LaCross Institute for Ethical AI in Business, Baugh Fellowship, the UVA Environmental Institute, MaceAI Trustworthy Compute Fellowship, OpenAI Researcher Program, and Google Gemini Award. The views expressed are those of the authors and do not reflect the official policy or the position of the funding agencies. N.Y. is supported, in part, by the SDS Pilot Grant and LaCross Institute for Ethical AI in Business.
\label{endbody}
\bibliography{radref}

\clearpage
\appendix

\onecolumn
\section{Worked example}
\label{apd:worked}

\figureref{fig:worked} shows a real reader-study case. Both reports are scored independently against the same image-derived beliefs. The radiologist report states a pleural effusion that the image beliefs support, while MAIRA-2
omits it. MAIRA-2 instead asserts mediastinal widening, for which the image belief is low. Both
reports deny pneumothorax, which the image strongly supports as absent. Across the complete finding
set, RadRef scores the radiologist report $+0.20$ and MAIRA-2 $+0.04$.

\begin{figure}[htbp]
\centering
\newcommand{\repblock}[3]{\colorbox{black!4}{\parbox{\dimexpr\textwidth-2\fboxsep\relax}{%
  \small\textbf{#1}#2\\[3pt]\normalsize\itshape #3}}}
\repblock{Radiologist report}{\hfill\footnotesize\color{black!55} shown only for comparison}{HISTORY: Worsening dyspnea, fever. TECHNIQUE: Frontal and lateral views of the chest. COMPARISON: \_\_\_. FINDINGS: Right-sided Port-A-Cath is again seen, terminating in the mid to distal SVC. The patient is status post median sternotomy. Right apical pleural thickening is again seen. Pleural thickening on the right lateral chest and right base/costophrenic angle again seen which \textbf{may be due to a small pleural effusion}. There has been interval development of patchy opacity at the lateral left lung base raising concern for developing pneumonia. There may be mild pulmonary vascular congestion. \textbf{No pneumothorax is seen}. The cardiac silhouette remains enlarged. IMPRESSION: Interval development of patchy opacity at the left lung base is concerning for pneumonia. Persistent elevation of the right hemidiaphragm and concern for right pleural effusion along with right pleural thickening. Mild pulmonary vascular congestion.}

\medskip
\repblock{MAIRA-2 report}{}{Right chest tube has been removed, with \textbf{no visible pneumothorax}. Multifocal post-radiation lung changes appear similar to the prior study, with a persistent peripheral right mid lung opacity corresponding to a site of previous radiation fibrosis. Cardiomediastinal contours are stable in appearance with \textbf{persistent widening of right mediastinal contours}, corresponding to a combination of post-radiation fibrosis and pre-existing mediastinal lymphadenopathy on prior PET-CT of \_. Marked gastric distention is noted within the imaged portion of the upper abdomen.}

\medskip
\noindent\small\textbf{Selected findings.} The generated report omits pleural effusion, asserts mediastinal widening, and correctly denies pneumothorax. The table shows how those claims contribute to the score.

\bigskip
{\small\begin{tabular}{@{}lcccc@{}}
\toprule
& & & & Contribution\\
Finding & Image belief & Radiologist & MAIRA-2 & (Radiologist / MAIRA-2)\\
\midrule
Pleural effusion$^{\dagger}$ & $.80$ & \textbf{present} & \textbf{omitted} & $+0.01$ / $0$\\
Mediastinal widening         & $.36$ & not mentioned & \textbf{present} & $0$ / $-0.02$\\
Pneumothorax                 & $.02$ & \textbf{absent} & \textbf{absent} & $+0.08$ / $+0.08$\\
\bottomrule
\end{tabular}}

{\footnotesize $^{\dagger}$ Omission identified by a radiologist reader.}

\bigskip
{\centering\rule{0.5\textwidth}{0.4pt}\\[4pt]
\small RadRef score\\[2pt]
\normalsize Radiologist $\;\mathbf{+0.20}$\qquad MAIRA-2 $\;\mathbf{+0.04}$\\[2pt]
\footnotesize higher is better\par}
\caption{Worked example. Both reports are scored against the same image-derived beliefs. The radiologist report is shown only for comparison. The table shows three finding-level contributions, while the final score uses the complete finding set.}
\label{fig:worked}
\end{figure}

\section{Evaluation details and diagnostics}
\label{apd:battery}
\label{apd:diagnostics}

\subsection{Paired ReXVal comparisons}
\label{apd:paired-rexval}
The main-text ReXVal table reports non-tied agreement. The following paired comparisons instead award half credit to metric ties and use shared study-level bootstrap resamples. The paired differences use the tie-inclusive statistic, while the main table reports agreement among non-tied metric pairs.

{\setlength{\tabcolsep}{5pt}
\begin{table}[htbp]
\floatconts
  {tab:paired}
  {\caption{Paired differences in ReXVal agreement between RadRef and each baseline. Ties count as one half. Positive values favor RadRef. Entries marked $^{*}$ have 95\% bootstrap CIs that exclude zero.}}
  {\footnotesize\begin{tabular}{@{}lrrr@{}}
  \toprule
  & \multicolumn{3}{c}{$\Delta$ RadRef $-$ baseline (95\% CI)}\\
  \cmidrule(lr){2-4}
  Metric & False findings & Omissions & Combined\\
  \midrule
  CRIMSON      & $-11.3^{*}$ ($-19.7$, $-2.3$) & $-3.0$ ($-14.2$, $9.6$) & $-8.3$ ($-17.3$, $1.4$)\\
  GREEN        & $-3.5$ ($-12.7$, $5.4$) & $-0.6$ ($-11.8$, $11.2$) & $-3.5$ ($-12.9$, $5.8$)\\
  RadFact      & $-6.1$ ($-13.5$, $1.0$) & $9.7^{*}$ ($0.9$, $19.1$) & $-0.9$ ($-9.3$, $7.9$)\\
  RadGraph-F1  & $-3.5$ ($-12.7$, $5.4$) & $14.2^{*}$ ($3.5$, $25.5$) & $0.2$ ($-9.4$, $8.9$)\\
  CheXbert-F1  & $7.8$ ($-0.5$, $16.5$) & $3.0$ ($-6.3$, $13.6$) & $7.8$ ($-1.9$, $18.0$)\\
  CheXbert-emb & $10.1^{*}$ ($0.5$, $20.6$) & $2.1$ ($-7.6$, $12.3$) & $7.1$ ($-3.4$, $18.6$)\\
  BERTScore    & $6.1$ ($-6.2$, $18.6$) & $8.8$ ($-3.6$, $21.0$) & $6.9$ ($-6.2$, $19.2$)\\
  ROUGE-L      & $15.8^{*}$ ($3.3$, $27.4$) & $12.1$ ($-2.2$, $26.4$) & $14.1^{*}$ ($0.2$, $26.2$)\\
  BLEU-4       & $19.6^{*}$ ($8.4$, $30.0$) & $14.8^{*}$ ($0.3$, $28.8$) & $19.1^{*}$ ($7.4$, $29.9$)\\
  \bottomrule
  \end{tabular}}
\end{table}}

\subsection{Empty-output control}

The cross-metric empty-output diagnostic uses MAIRA-2 on a common 150-study subset. The score-contrast analysis uses the archived 500-study cohort with per-report indication weighting. These analyses have different populations and weighting conventions. A defined empty-report RadRef score is zero; a report scores below it when negative finding-level contributions exceed positive credit.

\begin{table}[htbp]
\centering
\caption{Empty-report diagnostic on the 150-study common subset. Values award one point when the empty report outscores MAIRA-2 and one half for a tie, expressed as a percentage. For RadRef, this battery uses study-question weighting; it differs from the per-report weighting in the 500-study score-contrast analysis.}
\label{tab:empty-main}
\scriptsize
\setlength{\tabcolsep}{5pt}
\begin{tabular}{@{}lr@{}}
\toprule
Metric & Empty wins (\%)\\
\midrule
BLEU-4 & 1.0\\
BERTScore & 90.7\\
CheXbert-F1 & 21.0\\
CheXbert-emb & 13.3\\
RadGraph-F1 & 6.7\\
GREEN & 10.7\\
CRIMSON & 54.7\\
\midrule
RadRef (MedGemma-27B) & 30.7\\
\quad + calibration & 32.0\\
\quad + robust & 22.7\\
RadRef (Lingshu-32B) & 32.7\\
\quad + calibration & 33.3\\
\quad + robust & 15.3\\
\bottomrule
\end{tabular}
\end{table}

For a reference-based metric $M$, the diagnostic is $\Delta_{\mathrm{empty}}=M(G,R)-M(E,R)$;
for RadRef it is $\Delta_{\mathrm{empty}}=S(G,I)$, since $S(E,I)=0$. Here $G$ is the generated
report, $E$ the empty report, $R$ the radiologist reference and $I$ the radiograph. RadRef
therefore compares a generated report with silence without requiring a reference report.

\paragraph{Score ordering.}
A report's RadRef score measures its margin above or below the empty-report baseline. \tableref{tab:threedeltas} reports these margins on the archived 500-study cohort. Radiologist-authored reports have higher central scores than both generators, and both generators have positive mean and median scores; all reported confidence intervals exclude zero. This ordering describes the overall tendency rather than every study. In particular, $23.4\%$ of radiologist reports, $26.0\%$ of MAIRA-2 reports, and $19.6\%$ of LLaVA-Rad reports score below zero, reflecting negative contributions exceeding positive credit under the image-derived beliefs. Exact-zero counts are $3$, $23$, and $13$, respectively. 

\begin{table}[htbp]
\floatconts
  {tab:threedeltas}
  {\caption{RadRef score contrasts on the archived 500-study cohort under the primary configuration; this differs from the frozen 498-study preference cohort. Because an empty report scores exactly zero, the first three contrasts are the corresponding report scores. Intervals are $95\%$ study-level bootstrap confidence intervals. Positive (\%) is the percentage of studies for which the contrast is greater than zero.}}
  {\footnotesize\setlength{\tabcolsep}{4pt}\begin{tabular}{@{}lrrr@{}}
  \toprule
  Contrast & Median (95\% CI) & Mean (95\% CI) & Positive (\%)\\
  \midrule
  Radiologist $-$ empty & $+0.070$ ($+0.060$, $+0.083$) & $+0.055$ ($+0.039$, $+0.069$) & $76.0$\\
  LLaVA-Rad $-$ empty   & $+0.043$ ($+0.037$, $+0.047$) & $+0.036$ ($+0.029$, $+0.043$) & $77.8$\\
  MAIRA-2 $-$ empty     & $+0.037$ ($+0.030$, $+0.042$) & $+0.032$ ($+0.024$, $+0.040$) & $69.4$\\
  \midrule
  Radiologist $-$ LLaVA-Rad & $+0.029$ ($+0.020$, $+0.038$) & $+0.019$ ($+0.005$, $+0.033$) & $63.6$\\
  Radiologist $-$ MAIRA-2   & $+0.037$ ($+0.022$, $+0.048$) & $+0.023$ ($+0.008$, $+0.038$) & $65.8$\\
  \bottomrule
  \end{tabular}}
\end{table}

\section{Method details}

\subsection{Gate placement}
\label{sec:gate-placement}
The gate shrinks the numerator when image-derived beliefs are uncertain, moving the score toward the neutral baseline. Applying the same gate to the denominator produced identical same-radiograph rankings in every tested configuration. With the numerator-only gate, the maximum attainable score may be below $1$.

\subsection{Configuration parameters}
\label{apd:variantparams}
The alternative configurations use the same scoring rule as RadRef. \tableref{tab:variants} lists the changed parameters.

\paragraph{Calibration.} Some findings receive systematically higher or lower VLM beliefs than others, regardless of the particular radiograph. The $+$calibration configuration partially removes these finding-specific offsets. The implementation first fixes the most-specific finding inventory using the raw opposed-probe belief $I_n=(p_n+1-a_n)/2$. It estimates $L_n$, the mean logit of this raw belief over the cohort for finding $n$, and $G$, the mean of these finding-specific means. Findings require at least six observations; probabilities are clipped to $[10^{-4},1-10^{-4}]$ for logit operations. With $d_n=\alpha(L_n-G)$ and $\alpha=1/2$, it transforms the two complementary probe estimates separately: \[ p'_n=\sigma(\operatorname{logit}p_n-d_n),\qquad 1-a'_n=\sigma(\operatorname{logit}(1-a_n)-d_n). \] It then averages $p'_n$ and $1-a'_n$ and applies description-based fusion. A finding without an estimated offset is unchanged. A finding whose beliefs are systematically high is shifted downward, and one whose beliefs are systematically low is shifted upward, with half of the estimated offset removed. This centers the score on radiograph-specific departures from each finding's typical belief level, using neither reference reports nor reader annotations. The primary configuration omits it and needs no cohort statistics.

\paragraph{Robust configuration.} RadRef $+$robust softens the report forecasts, scoring assertions as $q^{+}=0.85$ and denials as $q^{-}=0.15$ in place of $1$ and $0$, and attenuates uncertain image beliefs more strongly ($\lambda = 4$). The parameter $\beta$ adds a probe-confidence weight $|2 I_n - 1|^{\beta}$, and $\varepsilon$ is an uncertainty band within which the mention rate $c_n$ replaces an unresolved probe belief. The robust configuration also includes the calibration correction. Its denominator retains the binary achievable-credit scale $s_n^{\star}$ from \equationref{eq:radref}, computed with forecasts $0$, $1/2$, and $1$.

\begin{table}[htbp]
\floatconts
  {tab:variants}
  {\caption{RadRef configurations.}}
  {\footnotesize\setlength{\tabcolsep}{2.6pt}\begin{tabular}{lccc}
  \toprule
   & RadRef & + calib. & + robust\\
  \midrule
  assertion forecast $q^{+}$ & $1.00$ & $1.00$ & $0.85$\\
  denial forecast $q^{-}$    & $0.00$ & $0.00$ & $0.15$\\
  gate strength $\lambda$       & $0.5$  & $0.5$  & $4$\\
  probe-belief confidence $\beta$    & $0$    & $0$    & $1$\\
  uncertainty band $\varepsilon$     & $0$    & $0$    & $0.1$\\
  belief-offset correction $\alpha$ & $0$    & $\tfrac12$ & $\tfrac12$\\
  \bottomrule
  \end{tabular}}
\end{table}

\subsection{Undefined scores and analysis eligibility}
\label{apd:undefined}
The scorer raises \texttt{EmptyDenominator} when there are no eligible findings or the achievable-credit denominator is nonpositive. Analysis wrappers record the result as missing or skip the affected report; they do not assign a zero score. A report comparison requires scores for both reports, and a paired metric comparison uses observations available for all quantities entering that comparison. Thus an undefined score is distinct from an empty report with a defined zero score.

\subsection{Calibration validity}
\label{sec:calibration-validity}
Leave-one-study-out recomputation of the belief-offset estimates changes no reported quantity by more than $0.2$ preference points, and the per-finding estimates correlate at $r=+0.976$ with an independent 500-study cohort. We fix $\alpha=\tfrac12$ a priori and report the uncalibrated configuration alongside it.

\section{Radiologist evaluation of model-generated reports}
\label{apd:reader}

\paragraph{Design.} Two radiologists independently reviewed 100 MIMIC-CXR studies disjoint from ReXVal, selected before annotation from the 500-study evaluation cohort (50 MAIRA-2 and 50 LLaVA-Rad generations).
\paragraph{Presentation.} For each study, readers saw the radiograph with the radiologist and generated reports in randomized, unlabeled positions. Both readers graded all 200 reports and were blinded to metric outputs.

{\setlength{\tabcolsep}{4pt}

\paragraph{Annotation protocol.} \tableref{tab:protocol} summarizes the reader instructions. Readers used the term \emph{overcall}. We use \emph{false finding} to match ReXVal terminology.

\begin{table}[htbp]
\floatconts
  {tab:protocol}
  {\caption{Reader-study annotation rules.}}
  {\footnotesize\begin{tabular}{@{}lp{0.70\linewidth}@{}}
  \toprule
  Item & Definition\\
  \midrule
  Task & Grade each report independently against the radiograph\\
  Omission & Finding visible on the radiograph but absent from the report\\
  False finding & Report claim unsupported by the radiograph, including calling an abnormal structure normal\\
  Attribute error & Wrong side, severity, or position, recorded as an omission with the error stated\\
  Ignored & Comparisons to prior studies, boilerplate, and anything beyond the findings themselves\\
  \bottomrule
  \end{tabular}}
\end{table}}

\paragraph{Same-radiograph ranking.} \tableref{tab:samerad} reports agreement on the
same-radiograph ranking, in which the radiologist's report and the generated report for a study
are ordered by omission count. Reference-based metrics cannot enter this comparison because the
radiologist report is their reference.

\begin{table}[htbp]
\floatconts
  {tab:samerad}
  {\caption{Same-image omission ranking across 100 cases using as-logged counts: one radiologist-versus-generated report pair per image. RadRef is compared with the two-reader mean; the inter-reader row compares the two readers on those within-image orders. Intervals are 95\% study-bootstrap intervals.}}
  {\footnotesize\setlength{\tabcolsep}{10pt}\begin{tabular}{@{}lrc@{}}
  \toprule
  Ranking compared with & $\tau_b$ & 95\% CI\\
  \midrule
  \method{}     & $.295$ & $.10$--$.47$\\
  Inter-reader  & $.386$ & $.20$--$.55$\\
  \bottomrule
  \end{tabular}}
\end{table}

\paragraph{\method{}-FF.} \method{}-FF (\sectionref{sec:ff}) isolates the false-finding evidence represented by the same finding-level beliefs. Assertion count, which uses no image information, serves as a length control.

\paragraph{Annotation targets and statistical methods.} The Kendall correlations and paired differences in \tableref{tab:samerad,tab:readerpaired} use as-logged counts, with the two-reader mean as their target. Main-text \tableref{tab:reader-main} instead uses protocol-audited counts, which reconcile marks with the annotation instructions, and reports each reader separately. We report Kendall's $\tau_b$ with 95\% intervals from 2{,}000 study-level bootstrap resamples shared across metrics. Pairwise comparisons are reported relative to the primary MedGemma RadRef configuration. The 100 studies come from 94 patients.

\paragraph{Annotation reliability.} The two-way random-effects ICC of the combined error count is $0.21$ (95\% CI $0.10$--$0.33$) for a single reader and $0.35$ ($0.17$--$0.50$) for the two-reader mean. Inter-reader $\tau_b$ measures how consistently the two radiologists rank the same reports by error count.

\paragraph{Metric comparison.} Main-text \tableref{tab:reader-main} reports each metric's agreement with each reader. We also compare each metric directly with RadRef using paired differences in Kendall's $\tau_b$ on shared bootstrap resamples (\tableref{tab:readerpaired}). For omissions, no compared metric differs significantly from RadRef.

{\setlength{\tabcolsep}{3pt}
\begin{table}[htbp]
\floatconts
  {tab:readerpaired}
  {\caption{Paired differences from RadRef across the 100 generated reports, as $\Delta\tau_b$ against the two-reader mean as-logged omission count on shared study-bootstrap resamples. These compare reports across images, unlike \tableref{tab:samerad}. Positive values favor the comparator. No interval excludes zero.}}
  {\scriptsize\begin{tabular}{@{}ll@{}}
  \toprule
  Metric & Omissions $\Delta\tau_b$ (95\% CI)\\
  \midrule
  CRIMSON       & $-0.027$ ($-0.21$, $0.14$)\\
  RadFact       & $-0.008$ ($-0.19$, $0.18$)\\
  GREEN         & $-0.029$ ($-0.23$, $0.17$)\\
  RadGraph-F1   & $-0.096$ ($-0.29$, $0.08$)\\
  CheXbert-F1   & $+0.056$ ($-0.11$, $+0.22$)\\
  CheXbert-emb  & $-0.002$ ($-0.17$, $+0.17$)\\
  BERTScore     & $-0.128$ ($-0.32$, $+0.05$)\\
  ROUGE-L       & $-0.052$ ($-0.25$, $+0.13$)\\
  BLEU-4        & $-0.030$ ($-0.21$, $+0.14$)\\
  \bottomrule
  \end{tabular}}
\end{table}}

\section{Sensitivity analyses and ablations}
\label{apd:extra}

\subsection{Image-probe validation}
\label{apd:probe-diagnostics}
Image-probe suitability is evaluated using finding-level discrimination, confident false-positive and false-negative rates, confidence coverage, and downstream agreement on a held-out radiologist-ranked set. The first three diagnostics here use asserted and explicitly denied findings extracted from radiologist reference reports, not independent image-only annotations. A belief is called confident when it is at least $0.75$ or at most $0.25$.

\begin{table}[htbp]
\centering
\caption{Probe quality and downstream ReXVal agreement. Confidence is $P(\hat I\leq.25\text{ or }\hat I\geq.75)$; FN and FP denote confident errors. Agreement excludes ties; parentheses are 95\% CIs.}
\label{tab:probe-diagnostics}
\scriptsize
\setlength{\tabcolsep}{3pt}
\begin{tabular}{@{}lrrrrr@{}}
\toprule
Probe & AUC & Conf. (\%) & FN (\%) & FP (\%) & Agreement (\%) \\
\midrule
MedGemma-27B & .824 & 55.2 & 17.5 & 4.2 & 73.1 (60.9--84.4) \\
Lingshu-32B & .696 & 36.2 & 9.5 & 12.5 & 68.6 (57.1--79.6) \\
RadVLM-7B & .878 & 6.5 & 0.0 & 0.0 & 60.0 (47.3--73.6) \\
MedGemma-4B & .587 & 37.8 & 0.0 & 45.8 & 43.3 (32.6--54.1) \\
\bottomrule
\end{tabular}
\end{table}
MedGemma-27B provides the highest held-out report-ranking agreement in these experiments while balancing discrimination, confident errors, and coverage. RadVLM-7B illustrates why a low confident-error rate alone is insufficient: it issues confident beliefs for just $6.5\%$ of evaluated findings. Updating the image-probe model may change RadRef scores and requires revalidation.

\paragraph{Probe dependence.}\label{apd:judge}
MedGemma-4B has negative rank association with ReXVal ($\tau_b=-0.103$). Its mean belief for findings asserted by the candidate but denied by the reference is $0.746$, compared with $0.302$ for MedGemma-27B and $0.402$ for Lingshu-32B, illustrating how confidently incorrect image beliefs affect false-finding evaluation.

\subsection{Component sensitivity}
\label{apd:component-sensitivity}
To assess sensitivity to individual pipeline choices, we change one component of the primary configuration at a time and measure non-tied combined agreement on ReXVal (\tableref{tab:component-sensitivity}).
\begin{table}[htbp]
\centering
\caption{Sensitivity to individual components. Non-tied ReXVal agreement and differences from primary RadRef (percentage points).}
\label{tab:component-sensitivity}
\footnotesize
\begin{tabular}{@{}lrr@{}}
\toprule
Setting & Agreement (\%) & Change \\
\midrule
Primary RadRef & 73.1 & --- \\
Description resampling (8 draws) & $72.9\pm2.1$ & $-0.2$ \\
SNOMED CT terminology & 69.3 & $-3.8$ \\
Lingshu-32B probe & 68.6 & $-4.5$ \\
$+$calibration & 76.1 & $+3.0$ \\
$+$robust & 70.1 & $-3.0$ \\
\bottomrule
\end{tabular}
\end{table}

\subsection{Analysis of below-empty scores}
\label{apd:below-empty}
\tableref{tab:below-empty-components} reports changes in below-empty rates under component substitutions. The MedGemma--Lingshu difference is $15.4$ percentage points; the tested parser alternatives change the rate by $-1.8$ to $0.0$ points, and vocabulary substitution changes it by $+0.2$ points. The below-empty rate varies more across the two image probes than across the tested parser alternatives or vocabularies.

\begin{table}[htbp]
\centering
\caption{Sensitivity of below-empty rates to pipeline components (percentage points). The probe row reports a between-model difference.}
\label{tab:below-empty-components}
\footnotesize
\setlength{\tabcolsep}{4pt}
\begin{tabular}{@{}lr@{}}
\toprule
Component changed & Change in below-empty rate (pp) \\
\midrule
Chest ImaGenome $\rightarrow$ SNOMED CT & $+0.2$ \\
Parser model (four alternatives) & $-1.8$ to $0.0$ \\
Deliberately poor parser prompt & $+7.2$ \\
Image probe (MedGemma vs. Lingshu) & $15.4$ difference \\
\bottomrule
\end{tabular}
\end{table}

\subsection{Free-form description stability}
\label{apd:description-stability}
Eight independent sets of five free-form descriptions were generated per radiograph on ReXVal, with all other components fixed. Variation across draws was modest relative to the bootstrap uncertainty in the primary results (\tableref{tab:description-resampling}).
\begin{table}[htbp]
\centering
\caption{Stability under resampling of free-form descriptions (eight draws on ReXVal). Agreement is in percent; mean $\pm$ SD.}
\label{tab:description-resampling}
\footnotesize
\begin{tabular}{@{}lrr@{}}
\toprule
Statistic & Reference draw & Mean $\pm$ SD \\
\midrule
RadRef $\tau_b$ & $+.353$ & $+.344\pm.031$ \\
Combined agreement & 73.13 & $72.87\pm2.07$ \\
False-finding agreement & 69.77 & $69.51\pm1.08$ \\
Omission agreement & 70.18 & $70.32\pm1.40$ \\
RadRef-FF $\tau_b$ & $+.324$ & $+.331\pm.016$ \\
RadRef-FF false-finding agreement & 86.67 & $87.14\pm0.83$ \\
\bottomrule
\end{tabular}
\end{table}

\subsection{Contribution of free-form descriptions}
\label{apd:sharedchan}

RadRef reads the radiograph through opposed probes and $M=5$ unprompted descriptions. RadVLM-7B \citep{deperrois2026radvlm} generates the descriptions, which are shared across the primary MedGemma and Lingshu configurations so that changing the probe model does not also change the description channel. In the $+$robust configuration, the mention rate can replace an unresolved probe belief. This fallback is used for about $1\%$ of findings.

Removing descriptions reduces radiologist preference under both standard probe configurations (\tableref{tab:sharedchan}).

\begin{table}[htbp]
\centering
\caption{Removing free-form descriptions. Changes in preference (percentage points) and empty-report win rate under the standard configuration.}
\label{tab:sharedchan}
\footnotesize
\setlength{\tabcolsep}{4pt}
\begin{tabular}{@{}lrrr@{}}
\toprule
Image probe & $\Delta$ MAIRA-2 pref. & $\Delta$ LLaVA-Rad pref. & $\Delta$ empty \\
\midrule
MedGemma-27B & $-5.2$ & $-3.8$ & $+8.4$ \\
Lingshu-32B & $-8.0$ & $-8.4$ & $+16.1$ \\
\bottomrule
\end{tabular}
\end{table}

\subsection{Finding-representation sensitivity}
\label{apd:kbsub}

\begin{table}[htbp]
\floatconts
  {tab:snomed}
  {\caption{Vocabulary substitution on ReXVal with the scoring rule unchanged.}}
  {\setlength{\tabcolsep}{2.5pt}\footnotesize\begin{tabular}{@{}lrrrr@{}}
  \toprule
  Inventory & Terms & Combined & Omissions & $\tau_b$\\
  \midrule
  Chest ImaGenome            & 69 & 73.1 & 70.2 & $+0.353$\\
  \quad grouped to 62        & 62 & 72.4 & 69.3 & $+0.342$\\
  SNOMED CT, re-probed       & 62 & 69.3 & 55.7 & $+0.299$\\
  \bottomrule
  \end{tabular}}
\end{table}

Keeping the scoring rule fixed, grouping Chest ImaGenome to 62 matched concepts changes combined agreement from $73.1\%$ to $72.4\%$. Re-probing the same radiographs with SNOMED CT terms gives $69.3\%$ (\tableref{tab:snomed}). The larger change on omissions ($70.2\%$ to $55.7\%$) shows that the scoring rule transfers across inventories, while the belief model remains sensitive to finding terminology.

\paragraph{Hierarchy de-duplication.}
\label{apd:dedup}

The standard formulation retains only the most specific radiograph-supported finding in each hierarchy chain, preventing one abnormality from contributing at multiple levels. Relative to scoring all hierarchy terms, this increases combined agreement from $68.6\%$ to $73.1\%$ and $\tau_b$ from $+0.305$ to $+0.353$, while reducing coverage from $70.5\%$ to $61.8\%$ (\tableref{tab:dedup}).

\begin{table}[htbp]
\floatconts
  {tab:dedup}
  {\caption{Effect of hierarchy de-duplication. Higher agreement and $\tau_b$ indicate better ranking. Coverage is the fraction of report pairs assigned distinct scores.}}
  {\setlength{\tabcolsep}{4pt}\footnotesize\begin{tabular}{@{}lrr@{}}
  \toprule
      & all terms & most-specific\\
  \midrule
  Within-study agreement         & 68.6 & 73.1\\
  Omissions                      & 67.4 & 70.2\\
  False findings                 & 66.7 & 69.8\\
  $\tau_b$                       & +0.305 & +0.353\\
  Coverage                       & 70.5 & 61.8\\
  \bottomrule
  \end{tabular}}
\end{table}

\subsection{Scoring-rule ablations}
\label{apd:ablate}
\paragraph{Component ablations.}
We remove one scoring component at a time while holding the radiograph beliefs and report parser
fixed (\tableref{tab:ablation}). RadRef makes two separate choices for unmentioned findings.
First, an unmentioned finding has its own forecast state, $R_n=\tfrac12$, rather than being
represented as asserted or denied. Second, the Brier score is centered at that forecast so an
unmentioned finding contributes exactly zero. The first two rows ablate these choices separately.
The table reports the percentage of comparisons in which the metric ranks the radiologist report
above MAIRA-2. A negative change means that removing the component lowers this preference rate.
For calibration of the magnitudes, an input-free assertion-count baseline reaches $56.8\%$ on
this target (\appendixref{apd:volctl}); \tableref{tab:degenerate} additionally reports the
alternative scoring rules on ReXVal agreement, where the same ordering holds.

\begin{table}[htbp]
\floatconts
  {tab:ablation}
  {\caption{One-at-a-time scoring ablations on the frozen 498-study cohort, using the saved cohort-pinned results. Preference awards one point when the radiologist report ranks above MAIRA-2 and one half for a tie. Changes are calculated before rounding.}}
  {\footnotesize\setlength{\tabcolsep}{3.2pt}\begin{tabular}{@{}lrr@{}}
  \toprule
  Ablation & Preference (\%) & Change\\
  \midrule
Full RadRef & 66.5 & ---\\
Remove distinct unmentioned state & 48.6 & $-17.9$\\
Remove zero-centering of unmentioned score & 52.0 & $-14.5$\\
Use binary rated/unrated relevance & 52.4 & $-14.1$\\
  Remove description-based belief floor & 59.8 & $-6.6$\\
Remove salience/confidence weighting & 62.2 & $-4.2$\\
Remove additional confidence gate & 63.5 & $-3.0$\\
Use weight-sum normalization & 66.3 & $-0.2$\\
  \bottomrule
  \end{tabular}}
\end{table}

The first ablation tests whether an unmentioned finding needs its own forecast state. The second keeps that state but tests whether it needs to be worth exactly zero. Both choices materially affect the preference rate, as does graded clinical relevance. The free-form description evidence also contributes, whereas per-radiograph normalization has little effect on this particular test. These preference analyses use report-specific indication weights. With identical weights and a shared positive denominator, zero-centering and normalization preserve within-study rankings. The description-floor ablation removes the floor; the description-channel analysis removes both the floor and mention-rate weighting. The normalization ablation divides by the sum of finding weights. The gate ablation retains confidence weighting.

\paragraph{Alternative scoring rules.}
\label{apd:degenerate}
We compare alternative scoring rules on ReXVal using the all-terms 69-finding inventory, with the matching all-terms RadRef result ($68.6\%$ combined agreement) as the comparator. Image labels are formed from the raw opposed-probe beliefs, before description-based belief flooring. Let $A=\{n\in\mathcal{K}:I_n\geq 1/2\}$ and let $B$ be the report's asserted findings within the probed inventory. Precision is $|A\cap B|/|B|$, recall is $|A\cap B|/|A|$, and $F_1$ is their harmonic mean. Accuracy is $(|\mathcal{K}|-|A\triangle B|)/|\mathcal{K}|$. A zero precision, recall, or harmonic-mean denominator returns zero. Explicit denials and unmentioned findings both lie outside $B$. These alternatives do not use relevance weights, description evidence, the Brier baseline, or the confidence gate.

\begin{table}[htbp]
\floatconts
  {tab:degenerate}
  {\caption{Alternative scoring rules on ReXVal using the all-terms inventory (50 studies, 200 reports). Agreement excludes metric ties. Length $\tau_b$ is association with report length.}}
  {\setlength{\tabcolsep}{5pt}\footnotesize\begin{tabular}{@{}lrrr@{}}
  \toprule
  Rule & Combined (\%) & Omissions (\%) & Length $\tau_b$\\
  \midrule
  $F_1$ & 50.9 & 61.5 & +0.296\\
  accuracy & 57.0 & 60.8 & -0.295\\
  precision & 58.4 & 62.8 & +0.244\\
  recall & 42.2 & 63.2 & +0.304\\
  \midrule
  RadRef, all terms & 68.6 & 67.4 & -0.067\\
  \bottomrule
  \end{tabular}}
\end{table}

The all-terms RadRef formulation has higher observed combined agreement than each tested alternative. This comparison evaluates complete scoring rules under the same inventory.

\paragraph{Parameter sensitivity.}
\label{apd:sensitivity}

We vary each robust-configuration parameter independently around its reported value:
$q^{-} \in [0.05, 0.25]$, $\varepsilon \in [0, 0.3]$, $\lambda \in [2, 6]$, and the
probe-belief confidence factor on or off. Radiologist preference varies by at most $3.2$
points under MedGemma-27B and $3.0$ under Lingshu-32B, the empty-report rate by at most $4.0$
points, and $\tau_b$ stays between $+0.283$ and $+0.330$. The robust configuration is
therefore locally stable to the individual parameter choices tested.

\subsection{Permutation scheme}
\label{apd:permscheme}
The permutation of \sectionref{sec:grounding} re-pairs reports and radiographs at random, so a
study can occasionally be scored against a radiograph whose findings resemble its own. We also
constructed pairings that maximize dissimilarity between the report's claims and the substituted
radiograph's image-derived beliefs, measured over the finding vocabulary and across normal and
abnormal studies. The larger effect under deliberately dissimilar pairings supports the same conclusion as the random re-pairing test.

\subsection{Clinical-indication sensitivity}
\label{apd:indctl}
We replace the stated indication with one sampled from another patient (random) or a maximally dissimilar indication (far), holding the radiograph, report, and beliefs fixed. \tableref{tab:indication} shows the primary configuration. Holding the remaining inputs fixed isolates the score's dependence on the clinical indication.

\begin{table}[htbp]
\centering
\caption{Indication substitution on 498 studies: primary RadRef radiologist-preference rates (\%). $\Delta$ compares stated versus random indication (95\% CI).}
\label{tab:indication}
\footnotesize
\setlength{\tabcolsep}{4pt}
\begin{tabular}{@{}lrrrc@{}}
\toprule
Comparison & Stated & Random & Far & $\Delta$ (95\% CI) \\
\midrule
vs MAIRA-2 & 66.5 & 63.8 & 30.3 & $+2.7$ ($-0.7$, $+6.0$) \\
vs LLaVA-Rad & 64.1 & 62.3 & 23.1 & $+1.7$ ($-1.7$, $+5.0$) \\
\bottomrule
\end{tabular}
\end{table}

\subsection{Simple baselines and length effects}
\label{apd:volctl}
\paragraph{Assertion count.}

An input-free baseline that prefers the report asserting more findings favors the radiologist on $56.8\%$ of studies against MAIRA-2 and $62.2\%$ against LLaVA-Rad, versus $66.5\%$ and $64.1\%$ for RadRef under the same tie-aware convention. Report volume therefore explains some, but not all, of the preference signal.

\paragraph{Report length.}
\label{apd:length}
\label{apd:pooled}

Pooling across studies is strongly confounded by report length: radiologist error count correlates with length at $\tau_b=+.546$. BLEU-4 rises from $45.1\%$ same-radiograph agreement to $70.7\%$ when studies are pooled, and every reference-based metric gains $7.1$--$25.6$ points. RadRef instead decreases by $9.3$ points. Its score--length correlation is $-0.099$.

\section{Prompts}
\label{apd:prompts}

Core RadRef prompts are reproduced below, with models and decoding settings in \tableref{tab:prompts}. The parser's auxiliary attribute menu is not used by RadRef scoring. Template variables appear in braces. Baseline implementations follow their released definitions where available. RadFact uses GPT-4o-mini in the reported runs.

\paragraph{Vocabulary linking.} Parsed finding names are linked to entries of $\mathcal{K}$ by a cached resolver: an exact match on the normalized name is tried first; otherwise a candidate menu of at most eight entries is assembled from token overlap and SapBERT embedding neighbors \citep{liu2021sapbert} with cosine similarity at least $0.40$, and GPT-4o-mini selects one menu entry or abstains (temperature $0$, decisions cached for determinism). A phrase that remains unlinked is excluded from scoring and contributes neither credit nor penalty.

\begin{table}[htbp]
\floatconts
  {tab:prompts}
  {\caption{Models, decoding settings, and outputs associated with each prompt.}}
  {\footnotesize\begin{tabular}{@{}llll@{}}
  \toprule
  Prompt & Model & Decoding & Produces\\
  \midrule
  Image probes        & MedGemma-27B & greedy, $2\times$69 per radiograph & $p_n$, $a_n$\\
  Unprompted descriptions & RadVLM-7B & $T=1.0$, $M=5$ per radiograph   & $c_n$\\
  Report parsing & GPT-4o-mini & greedy, once per report & $R_n$\\
  Relevance tiers     & GPT-4o-mini  & once per indication           & $t_n$\\
  \midrule
  Report generation     & MAIRA-2 / LLaVA-Rad & public model settings & candidate reports\\
  \bottomrule
  \end{tabular}}
\end{table}

\subsection{Image probes}
\label{apd:pr:probe}
Opposed image probes estimate $p_n$ and $a_n$. First-token probabilities are renormalized over \{Yes, No\}.
\begin{promptbox}[MedGemma-27B, greedy]{Image probes}
Is \{finding\} present in the chest X-ray?\\
Answer Yes or No.

Is the chest X-ray free of \{finding\}?\\
Answer Yes or No.
\end{promptbox}

\subsection{Free-form image description}
\label{apd:pr:desc}
Used for the description-based recall correction of \equationref{eq:descfloor}.
\begin{promptbox}[RadVLM-7B, $T = 1.0$, $M = 5$]{Unprompted image descriptions}
Describe the findings in this chest X-ray.
\end{promptbox}

\subsection{Report parsing}
\label{apd:pr:parsesys}
\label{apd:pr:parse}
System message:
\begin{promptbox}[system message]{Report parsing}
You are a chest radiologist extracting findings and their described attributes from a report, for
structured comparison against the image.
\end{promptbox}

User message (the auxiliary attribute menu is not used for RadRef scoring):
\begin{promptbox}[user message]{Report parsing}
Extract every finding the report asserts as PRESENT (including hedged mentions like \pq{}possible
effusion\pq{}) or explicitly ABSENT/RESOLVED (\pq{}no effusion\pq{}, \pq{}lungs are clear\pq{},
\pq{}resolved\pq{}). Include support devices (ng/et/feeding tubes, central/PICC/IJ lines, chest
tubes, pacemakers) named by the device.

For each finding return:

\pq{}name\pq{}: a canonical finding name (map synonyms to one name: volume loss -> atelectasis;
infiltrate -> consolidation; vascular congestion -> pulmonary edema; enlarged heart ->
cardiomegaly). Keep the SPECIFIC entity when the report is specific (e.g. \pq{}cavitary
lesion\pq{}, not \pq{}opacity\pq{}).

\pq{}present\pq{}: true if asserted present/possible, false if asserted absent/resolved.

\pq{}axes\pq{}: for each axis the report SPECIFIES, output EXACTLY ONE word from that axis's
allowed set. Omit any axis the report does not give. \{axis\_menu\} is filled from the parser attribute menu, which is not used in RadRef scoring.

Return JSON only:\\
\{\pq{}findings\pq{}: [\{\pq{}name\pq{}: \pq{}...\pq{}, \pq{}present\pq{}: true,
\pq{}axes\pq{}: \{\pq{}location\pq{}: \pq{}lower\pq{}\}\}]\}

Report:\\
\{text\}
\end{promptbox}

\subsection{Relevance weighting}
\label{apd:pr:tiers}
The five tiers map to $t_n \in \{1.0, 0.7, 0.5, 0.3, 0.1\}$.
\begin{promptbox}[one call per indication]{Clinical relevance tiers}
Given the clinical indication, assign each finding to exactly one tier:

\pind directly-assessed = the finding IS what the exam was ordered to evaluate\\
\pind same-system = same organ system as the question\\
\pind moderately-relevant\\
\pind incidental\\
\pind unrelated

Return JSON only: \{\pq{}finding\pq{}: \pq{}tier\pq{}, ...\}
\end{promptbox}

\subsection{Generation prompt used for evaluation data}
\label{apd:pr:gen}
MAIRA-2 uses the authors' published image-only generation procedure.
\begin{promptbox}[LLaVA-Rad, public model]{Report generation}
Describe the findings of the chest x-ray.
\end{promptbox}

\end{document}